%% file: main.tex
\documentclass[conference]{IEEEtran}
\IEEEoverridecommandlockouts

\usepackage{float}
\usepackage{amsmath}  
\usepackage{amssymb}
\usepackage{graphicx}  
\usepackage{xcolor}
\usepackage{booktabs}  
\usepackage{subcaption}
\usepackage[capitalise]{cleveref}  
\usepackage{makecell}  
\usepackage[nolist,nohyperlinks]{acronym}  
\usepackage{siunitx}
\usepackage{censor}
\usepackage{cite}

\newcommand{\algname}{HumanHalo}

\input{notation}

\acrodef{MAV}[MAV]{Micro Air Vehicle}
\acrodef{POMDP}[POMDP]{Partially Observable Markov Decision Process}
\acrodef{ORCA}[ORCA]{Optimal Reciprocal Collision Avoidance}
\acrodef{HJ}[HJ]{Hamilton--Jacobi}
\acrodef{RL}[RL]{Reinforcement Learning}
\acrodef{DOF}[DOF]{Degrees of Freedom}
\acrodef{MPC}[MPC]{Model Predictive Control}
\acrodef{CBF}[CBF]{Control Barrier Function}
\acrodef{QP}[QP]{Quadratic Program}

\title{HumanHalo: Safe and Efficient 3D Navigation Among Humans via Minimally Conservative MPC}

\author{
\IEEEauthorblockN{
Simon Schaefer\IEEEauthorrefmark{1}\IEEEauthorrefmark{2},
Helen Oleynikova\IEEEauthorrefmark{2},
Sandra Hirche\IEEEauthorrefmark{1},
and Stefan Leutenegger\IEEEauthorrefmark{2}
}
\IEEEauthorblockA{
\IEEEauthorrefmark{1}Technical University of Munich, Germany\\
\IEEEauthorrefmark{2}ETH Zurich, Switzerland
}
}

\begin{document}

\maketitle
\thispagestyle{empty}
\pagestyle{empty}

\begin{abstract}
Safe and efficient robotic navigation among humans is essential for integrating robots into everyday environments. Most existing approaches focus on simplified 2D crowd navigation and fail to account for the full complexity of human body dynamics beyond root motion. 
We present \algname, an \ac{MPC} framework for 3D \ac{MAV} navigation among humans that combines theoretical safety guarantees with data-driven models for realistic human motion forecasting. Our approach introduces a novel reachability-based safety formulation that constrains only the initial control input for safety while modeling its effects over the entire planning horizon, enabling safe yet efficient navigation.
We validate \algname{} in both simulated experiments using real human trajectories and in the real world, demonstrating its effectiveness across tasks ranging from goal-directed navigation to visual servoing for human tracking. While we apply our method to \acp{MAV} in this work, it is generic and can be adapted to other platforms. Our results show that the method preserves safety without excessive conservatism, even under imperfect human prediction and in constrained workspaces, while remaining efficient enough for real-time onboard execution.
Our code will be made publicly available upon acceptance.
\end{abstract}


\input{sections/01_introduction}
\input{sections/02_related_work}
\input{sections/03_preliminaries}
\input{sections/05_approach}
\input{sections/06_experiments}
\input{sections/07_conclusion}


\bibliographystyle{IEEEtran}
\bibliography{references}

\end{document}

%% file: notation.tex
\usepackage{xifthen}
\usepackage{xparse}
\usepackage{subdepth}
\usepackage{leftidx}
\usepackage{bm}

\newcommand{\bbm}{\begin{bmatrix}}
\newcommand{\ebm}{\end{bmatrix}}

\DeclareMathAlphabet{\mbf}{OT1}{ptm}{b}{n}
\newcommand{\mbs}[1]{{\bm{#1}}} 

\newcommand{\mbsbar}[1]{{\overline{\boldsymbol{#1}}}}
\newcommand{\mbshat}[1]{{\hat{\boldsymbol{#1}}}}
\newcommand{\mbstilde}[1]{{\tilde{\boldsymbol{#1}}}}

\newcommand{\mbsdot}[1]{{\dot {\boldsymbol{#1}}}}

\newcommand{\mbfbar}[1]{{\overline{\mbf{#1}}}}
\newcommand{\mbfhat}[1]{{\hat{\mbf{#1}}}}
\newcommand{\mbftilde}[1]{{\tilde{\mbf{#1}}}}

\newcommand{\mbfdot}[1]{{\dot{\mbf{#1}}}}

\newcommand{\cframe}[1]{{\smash{\protect\underrightarrow{\mathcal{F}}_{#1}}}}

\DeclareMathAlphabet{\mathbfit}{OML}{cmm}{b}{it}
\newcommand{\homo}[1]{{\mathbfit{#1}}}

\newcommand{\mbfh}[1]{{\homo{#1}}}

\newcommand{\pos}[2]{\leftidx{_{#1}}{ \mbf r}{_{#2}}} 
\newcommand{\posh}[2]{\leftidx{_{#1}}{\mbfh r}{_{#2}}} 

\newcommand{\vel}[3]{\leftidx{_{#1}}{\mbf v}{\IfValueTF{#2}{_{#2#3\hspace{2pt}}}{}}} 
\newcommand{\veltilde}[3]{\leftidx{_{#1}}{\mbftilde v}{\IfValueTF{#2}{_{#2#3\hspace{2pt}}}{}}} 
\newcommand{\velbar}[3]{\leftidx{_{#1}}{\mbfbar v}{\IfValueTF{#2}{_{#2#3\hspace{2pt}}}{}}} 
\newcommand{\velhat}[3]{\leftidx{_{#1}}{\mbfhat v}{\IfValueTF{#2}{_{#2#3\hspace{2pt}}}{}}} 
\newcommand{\veldot}[3]{\leftidx{_{#1}}{\mbfdot v}{\IfValueTF{#2}{_{#2#3\hspace{2pt}}}{}}} 

\newcommand{\acc}[3]{\leftidx{_{#1}}{\mbf a}{\IfValueTF{#2}{_{#2#3\hspace{2pt}}}{}}} 
\newcommand{\acctilde}[3]{\leftidx{_{#1}}{\mbftilde a}{\IfValueTF{#2}{_{#2#3\hspace{2pt}}}{}}} 
\newcommand{\accbar}[3]{\leftidx{_{#1}}{\mbfbar a}{\IfValueTF{#2}{_{#2#3\hspace{2pt}}}{}}} 

\newcommand{\rotvel}[3]{\leftidx{_{#1}}{\mbs \omega}{\IfValueTF{#2}{_{#2#3\hspace{2pt}}}{}}} 
\newcommand{\rotveltilde}[3]{\leftidx{_{#1}}{\mbstilde \omega}{\IfValueTF{#2}{_{#2#3\hspace{2pt}}}{}}} 
\newcommand{\rotvelbar}[3]{\leftidx{_{#1}}{\mbsbar \omega}{\IfValueTF{#2}{_{#2#3\hspace{2pt}}}{}}} 
\newcommand{\rotvelhat}[3]{\leftidx{_{#1}}{\mbshat \omega}{\IfValueTF{#2}{_{#2#3\hspace{2pt}}}{}}} 
\newcommand{\rotveldot}[3]{\leftidx{_{#1}}{\mbsdot \omega}{\IfValueTF{#2}{_{#2#3\hspace{2pt}}}{}}} 

\newcommand{\T}[2]{\leftidx{}{\mbfh T}{_{#1#2\hspace{2pt}}}} 
\newcommand{\q}[3]{\leftidx{_{#3}}{\mbf q}{_{#1#2\hspace{2pt}}}} 


%% file: sections/01_introduction.tex
\section{Introduction}
Safely and efficiently controlling robots in dynamic environments remains a fundamental challenge. Existing approaches either rely on end-to-end learned policies to handle inherent uncertainties~\cite{cheng2019endtoend}, probabilistic frameworks with simplified dynamics~\cite{mohamed2025chanceconstrained}, or reachability-based methods that provide formal guarantees but are often overly conservative or computationally prohibitive for complex environments and resource-constrained platforms~\cite{thumm2025generalsafetyframework}.
To address these limitations, we propose a novel control framework for efficient and safe navigation that builds on reachability-based methods while introducing a conceptual simplification: for MPC-like formulations it suffices to enforce safety only on the initial control input while accounting for its effect over the entire planning horizon. This removes complex nonlinear couplings, enabling computationally efficient solutions while still providing recursive safety guarantees at every executed control step.

\begin{figure}[t!]
\centering
\includegraphics[width=\linewidth]{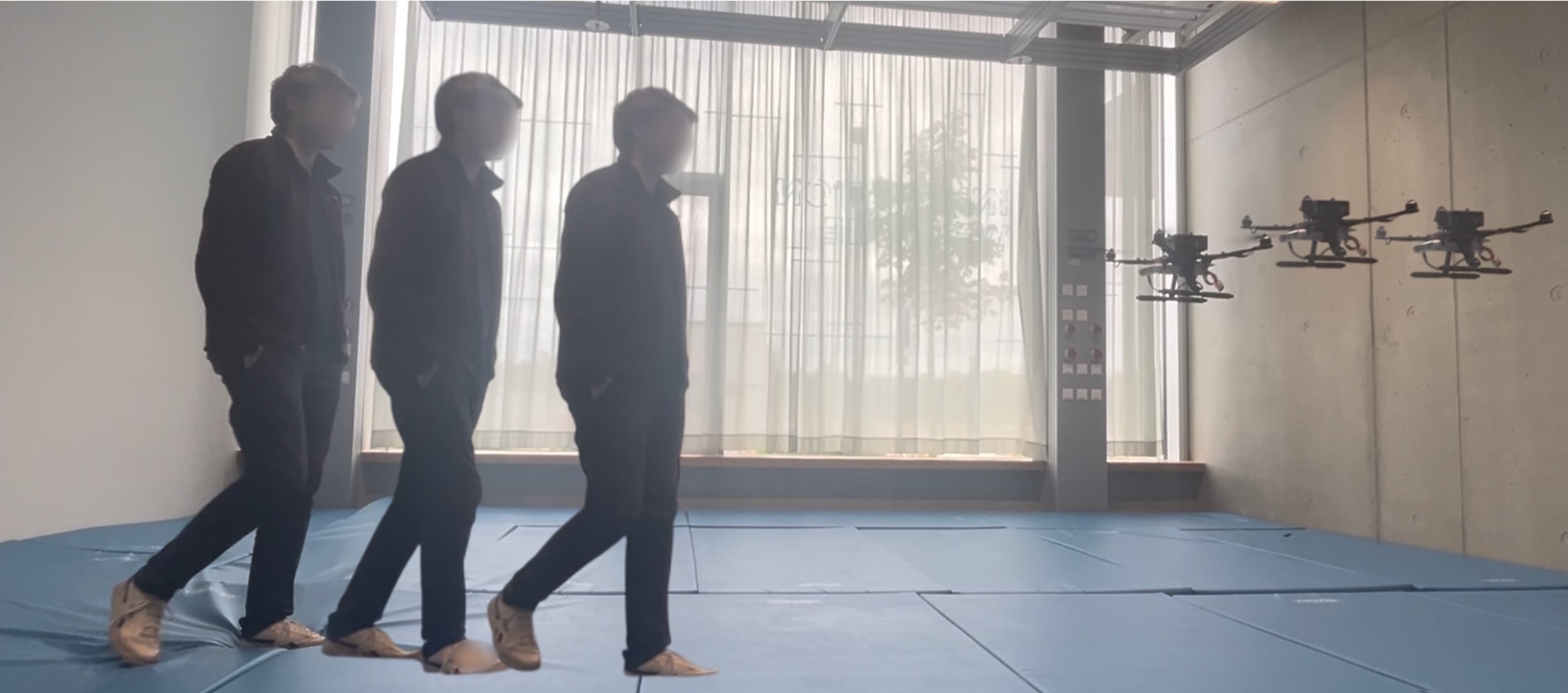}
\caption{\algname: A control framework for safe and efficient MAV navigation among humans. Our method combines reachability-based safety constraints with state-of-the-art data-driven human motion forecasting, ensuring recursive safety guarantees without overly conservative behavior.}
\label{fig:real_world_experiments}
\vspace{-0.6cm}
\end{figure}

We demonstrate the proposed framework on navigation in human-populated environments, as illustrated in \cref{fig:real_world_experiments}, where the dynamic, uncertain, and full-body nature of human motion makes collision-free flight particularly challenging. Most prior work targets 2D crowd settings that model humans as moving points in a plane~\cite{trautman2015robotcrowdnavigation, sun2021trajectories, bai2015intentionaware, dragan2017mathematicalmodels, samavi2025sicnav}, a severe simplification for robots operating in close proximity, which must reason about full-body motion in three dimensions. Recent 3D efforts remain limited, e.g.\ to stationary humans~\cite{salazar2024hierarchical}, while learning-based methods~\cite{cheng2019endtoend, everett2021collision, kretzschmar2016sociallycompliant} and Safe \ac{RL}~\cite{garcia2015safeRL} offer only probabilistic, in-distribution safety at a high, hard-to-collect data cost.

Building on our new reachability formulation, we introduce \algname, a control framework for safe and efficient \ac{MAV} navigation in 3D human environments. Unlike prior approaches, our method provides safety without restrictive simplifications, excessive conservatism, or extensive precomputation. It is sufficiently efficient to explicitly reason about full 3D human body motion while remaining feasible for onboard execution on an \ac{MAV}. Furthermore, it integrates state-of-the-art data-driven human motion forecasting rather than relying on handcrafted motion models. In summary, the main contributions of this work are as follows:

\begin{itemize}
\item We introduce a new safety constraint for MPC that provides a formal recursive safety guarantee under stated reachability, model, and recursive-feasibility assumptions, while remaining linear and real-time solvable. In contrast to \ac{HJ} reachability, our formulation avoids extensive precomputation and model simplifications while retaining the same linear, per-step structure as forward reachability. This enables scalable and practical integration into online optimization.
\item We design an MPC framework for safe and efficient \ac{MAV} navigation among humans. By combining our safety constraint with state-of-the-art human motion forecasting, the method avoids overly conservative behavior and simplistic motion assumptions. It leverages nominally optimistic human motion estimates while enforcing as-safe-as-necessary assurances, and integrates static obstacles as linear workspace constraints. This results in a computationally efficient QP suitable for real-time onboard deployment.
\item We validate the approach against external and custom baselines in simulation using real human motion trajectories and in real-world experiments on an \ac{MAV}. The results demonstrate effective and versatile performance across tasks ranging from goal-directed navigation to visual servoing for human tracking, and show that, unlike barrier- and prediction-based baselines, our method stays collision-free under imperfect prediction without overly conservative navigation.
\end{itemize}


%% file: sections/02_related_work.tex
\section{Related Work}
Safe navigation among humans has long been studied for ground robots and, more recently, aerial platforms, where a recurring challenge is balancing formal safety guarantees against realistic human-motion modeling and computational tractability. One line of work models human cooperation and intent to avoid overly conservative, ``freezing'' behavior. Trautman et al.~\cite{trautman2015robotcrowdnavigation} introduced interacting Gaussian processes to capture cooperative collision avoidance, showing that explicitly modeling cooperation avoids the freezing-robot problem and rivals human teleoperators in dense crowds. Sun et al.~\cite{sun2021trajectories} reason in the space of preference distributions rather than trajectories, richly representing human willingness to cooperate while retaining real-time performance. Bai et al.~\cite{bai2015intentionaware} cast pedestrian interaction as a \ac{POMDP} to hedge against intent uncertainty, and Dragan~\cite{dragan2017mathematicalmodels} grounds navigation in game-theoretic models of human cognition. These achieve strong crowd performance but rely on simplified planar trajectory models and can struggle in highly dynamic, multi-human settings.

Safety-critical methods based on \ac{HJ} reachability~\cite{mitchell2005hjreachability, fridovichkeil2018planningfastslow} provide formal guarantees through precomputed trajectory tubes but scale poorly, forcing overly simplistic dynamics. \acp{CBF} are an efficient alternative that filter a nominal controller~\cite{ames2017cbf} or augment an MPC~\cite{zeng2021mpccbf} with a barrier constraint. Robust variants absorb bounded uncertainty, but their distance-margin barrier forbids overlap and grows conservative as the set inflates, whereas we forbid only containment. Stefanini et al.~\cite{stefanini2024efficientcontext} integrate 3D skeletal and activity cues into an efficient context-aware MPC, but project them into planar avoidance constraints without formal safety guarantees. \ac{ORCA}-based approaches~\cite{vandenberg2011orca, samavi2025sicnav, samavi2025sicnavdiffusion} give efficient multi-agent avoidance under simplified velocity-obstacle models, and hybrid methods pairing learned forecasting with \ac{HJ} safety~\cite{schaefer2020leveragingnngradients} improve realism but remain costly and largely 2D.

Learning-based approaches, including \ac{RL} and inverse \ac{RL}~\cite{cheng2019endtoend, everett2021collision, kretzschmar2016sociallycompliant}, capture implicit interaction patterns and socially compliant behavior. Safe \ac{RL}~\cite{garcia2015safeRL} and, more recently, foundation-model policies further improve robustness, yet their safety assurances remain largely probabilistic and valid mainly in-distribution, and they are sample-inefficient. Applications to aerial robots~\cite{tallamraju2020aircaprl} explore drone-based human tracking but cannot operate robustly in close proximity, and recent hierarchical drone navigation among humans~\cite{salazar2024hierarchical} is limited to stationary humans and static obstacles.

In summary, prior methods trade formal guarantees against conservative navigation, richer human models, or scalability. We instead combine reachability-based safety with data-driven human motion forecasting for real-time, efficient, and, under stated assumptions, provably safe navigation of aerial robots among humans.

%% file: sections/03_preliminaries.tex
\begin{figure*}[ht!]
    \centering
    \includegraphics[width=\linewidth]{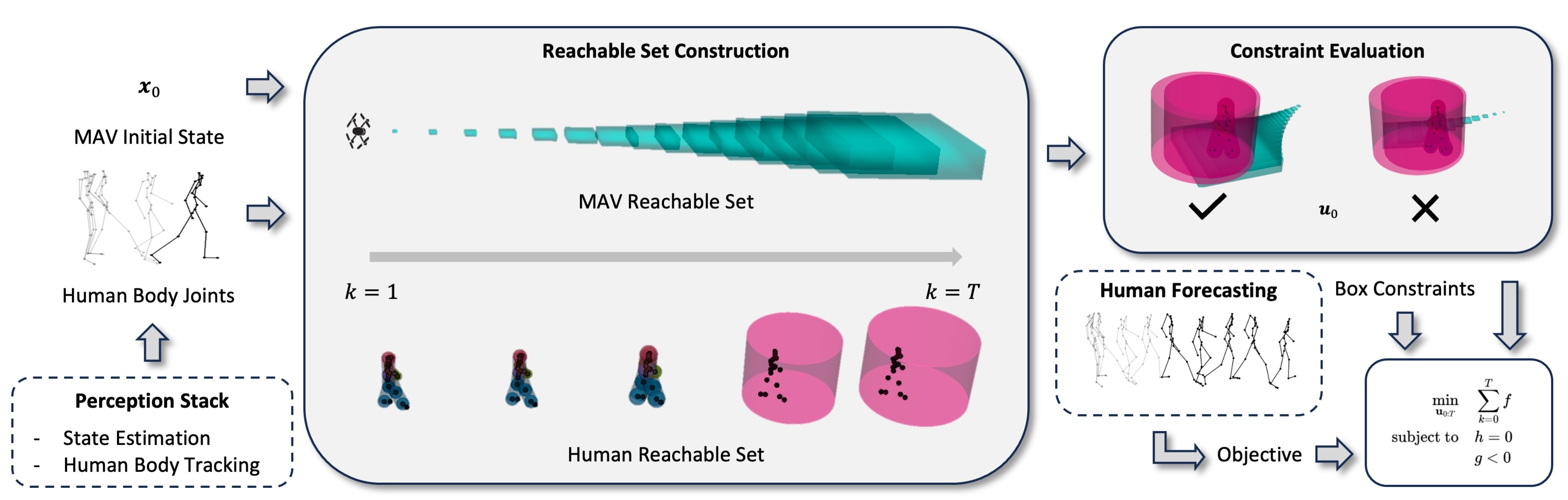}
    \caption{Based on the current \ac{MAV} state $\mathbf{x}_0$ and the tracked 3D positions of human body joints, we first compute both the \ac{MAV}'s and the human's reachable sets over the entire horizon. We then optimize the control inputs $\mathbf{u}_k$ with the requirement that the initial control input $\mathbf{u}_0$ must not lead to a situation where the \ac{MAV}’s reachable set becomes a subset of the human’s reachable set at any time. While partial overlap between the two sets is safe, as it can be resolved by future control actions, full \emph{containment} of the \ac{MAV}'s set within the human's would imply inevitable collision regardless of future control inputs. To run on an \ac{MAV}, our method is augmented with a real-time perception and forecasting stack (dashed boxes) that provides the current human and \ac{MAV} state and informs the objective about future human motion.}
    \label{fig:system_overview}
    \vspace{-0.4cm}
\end{figure*}

\section{Preliminaries}
\label{sec:preliminaries}
\subsection{Coordinate Frames, Transformations, and Notation}
Reference coordinate frames are denoted $\cframe{A}$, points $\pos{A}{}$ with homogeneous form $\posh{A}{}$, and the transform from $\cframe{B}$ to $\cframe{A}$ is $\T{A}{B}$ (position $\pos{A}{A B}$, orientation $\q{A}{B}{}$). We use the world frame $\cframe{W}$, human-centric frame $\cframe{H}$, and \ac{MAV} body frame $\cframe{N}$. The human body is the 24 SMPL joints~\cite{loper2015smpl} in the world frame, ${_W}\mathbf{J}^H$.

\subsection{MAV Model}
The \ac{MAV} state contains its world-frame position $\pos{W}{B}$, orientation $\q{W}{B}{}$ as ZYX Euler angles (yaw $\psi$, pitch $\theta$, roll $\phi$)~\cite{tzoumanikas2019linearmpc}, and linear velocity $\vel{W}{}{}$:
\begin{align}
\mathbf{x} = [\pos{W}{B}, \q{W}{B}{}, \vel{W}{}{}].
\label{eq:state_definition}
\end{align}
The navigation frame $\cframe{N}$ used for control rotates $\cframe{W}$ about $z$ by $\psi$ and translates to the \ac{MAV} position, $\posh{N}{} = \T{N}{B} \posh{B}{}$. Following \cite{tzoumanikas2019linearmpc, darivianakis2014hybrid}, the dynamics are a second-order system:
\begin{subequations}
\begin{alignat}{2}
\ddot{x} &= g \theta - c_x \dot{x}, \\
\ddot{y} &= - g \phi - c_y \dot{y}, \\
\ddot{z} &= \tau - c_z \dot{z}, \\
\ddot{\theta} &= - b_1 \theta + b_2 \theta^r, \\
\ddot{\phi} &= - b_3 \phi + b_4 \phi^r,
\end{alignat}
\label{eq:model}
\end{subequations}
with inputs $\mathbf{\mathit{u}_t} = [\tau, \theta^r, \phi^r]$, drag coefficients $c_i$, and gravity $g$. The $c_i$ and $b_i$ are system-identified. Exploiting differential flatness w.r.t.\ yaw~\cite{tzoumanikas2019linearmpc} and assuming isotropy, small attitude angles, and negligible aerodynamic coupling, the dynamics reduce to the linear system \eqref{eq:model}. This holds in the near-hover, moderate-speed regime typical of close human proximity. Aggressive high-speed flight would require the nonlinear model. Yaw is regulated separately and is not part of the linear control model~\eqref{eq:model}, with its rate command tracked fast enough to face the observed human.
With a sampling time $T_s$, \eqref{eq:model} yields the discrete, time-invariant state-space form
\begin{align}
\mathbf{x}_{k+1} &= \mathbf{A} \mathbf{x}_k + \mathbf{B} \mathbf{u}_k, \\
\mathbf{x}_{0:T} &= \mathbf{\Phi} \mathbf{x}_0 + \mathbf{\Gamma} \mathbf{u}_{0:T-1},
\label{eq:system_dynamics}
\end{align}
where $\mathbf{A}, \mathbf{B}$ are the state-space matrices and $\mathbf{\Phi}, \mathbf{\Gamma}$ their stacked forms.

%% file: sections/05_approach.tex
\section{Safe Control among Humans}
\label{sec:approach}
We formulate a finite-horizon linear \ac{MPC} problem that jointly optimizes task-specific objectives while enforcing safety with respect to nearby humans (see \cref{fig:system_overview}). The optimization problem is

\begin{subequations}
\begin{alignat}{2}
\min_{\mathbf{u}_{0:T-1}} \quad & \sum_{k=0}^T l_k^{\mathrm{TR}} \\
\text{subject to} \quad & \mathbf{x}_{k+1} = \mathbf{A} \mathbf{x}_k + \mathbf{B} \mathbf{u}_k, \quad k=0,\ldots,T-1, \\
& \mathbf{u}_{\min} \leq \mathbf{u}_k \leq \mathbf{u}_{\max}, \\
& \mathbf{x}_{\min} \leq \mathbf{x}_k \leq \mathbf{x}_{\max}, \\
& h_i\big(\mathbf{u}_0, {_W}\mathbf{J}^H_k \big) \geq 0, \quad \forall i, k, \\
& f_j(\mathbf{x}_k) \geq 0, \quad \forall j, k.
\end{alignat}
\label{eq:optimization_problem}
\end{subequations}
Here, $\mathbf{x}_k \in \mathbb{R}^n$ and $\mathbf{u}_k \in \mathbb{R}^m$ are the state and control input at step $k$. A key feature of this formulation is that the objective $l_k^{\mathrm{TR}}$ may freely use any human-motion forecast, however complex, such as the predicted future joints ${_W}\mathbf{J}^H_{k+1:k+T}$, which are often accurate but inherently error-prone. Safety, by contrast, rests on the hard constraints $h_i(\cdot)$, which use only the currently observed joints ${_W}\mathbf{J}^H_k$ and thus need neither a forecast nor a simplified human motion assumption, yet remain safe without being overly conservative. Each human contributes several convex body primitives, indexed by $i$ (\cref{sec:human_reachable_set}), so the problem grows linearly in the number of people, while the functions $f_j(\cdot)$ keep the \ac{MAV} within a static workspace. The formulation is embodiment-independent.

\subsection{Tracking Objective}
\label{sec:tracking_objective}
In this work, we focus on two tasks, goal-directed setpoint navigation and visual servoing for human body tracking. While the first involves a simple $\ell_2$ tracking error between the states $\mathbf{x}_k$ and a goal state, the latter uses the body motion forecasting model to control the \ac{MAV} to maintain a constant offset to the front of the human body. Based on the forecasted human joints ${_W}\mathbf{J}^H_{k+1:k+T}$, a reference trajectory is computed and tracked using the following objective function:
\begin{equation}
l_k^{\mathrm{TR}} = ||\mathbf{x}_k - \hat{\mathbf{x}}_k(_{W}\mathbf{J}^H_{k+1:k+T})||^2.
\end{equation}

\subsection{Reachability Constraint}
\label{sec:reachability_constraint}
Let $\mathbf{u}_0$ be the initial control input executed, $\mathcal{R}^R_k(\mathbf{u}_0)$ the \ac{MAV} reachable set at time $k$ under $\mathbf{u}_0$, and $\mathcal{R}^H_{k}$ the human body’s reachable set at time $k$ over the planning horizon $[0,T]$. The reachability constraint (RC) requires that for every $k\in[1,T]$

\begin{equation}
\mathcal{R}^R_k(\mathbf{u}_0)\not\subseteq \mathcal{R}^H_{k}.    
\end{equation}
If this holds, some reachable position remains outside the human set. If instead $\mathcal{R}^R_k(\mathbf{u}_0)\subseteq \mathcal{R}^H_{k}$, no reachable position is collision-free against every admissible human motion, an inevitable-collision state~\cite{fraichard2004ics}. Enforcing non-containment therefore avoids these states, and the recursive argument of \cref{sec:safety_analysis} under Assumption 3 turns this into closed-loop safety.

We measure containment with a signed distance $d(\mathcal S_1,\mathcal S_2)$ via the boundary point $\mathbf p \in \partial \mathcal S_2$ closest to $\partial \mathcal S_1$, negative (the maximum distance into $\mathcal S_1$) if $\mathcal S_1 \subseteq \mathcal S_2$ and positive otherwise,

\begin{equation}
d(\mathcal S_1,\mathcal S_2)
=
\begin{cases}
-\displaystyle \min_{\mathbf s_1 \in \mathcal S_1} \lVert \mathbf s_1 - \mathbf p \rVert, & \mathcal S_1 \subseteq \mathcal S_2,\\[2mm]
\displaystyle \max_{\mathbf s_1 \in \mathcal S_1} \lVert \mathbf s_1 - \mathbf p \rVert, & \text{otherwise.}
\end{cases}
\end{equation}

The distance per set $i$ must then be greater than zero:
\begin{equation}
h_i(\bm{J}^H_{k}, \mathbf{u}_0) = d(\mathcal{R}^R_k, \mathcal{R}^H_{k, i}) > 0.
\label{eq:reachability_general}
\end{equation}

\begin{figure}[ht!]
\centering
\includegraphics[width=\linewidth]{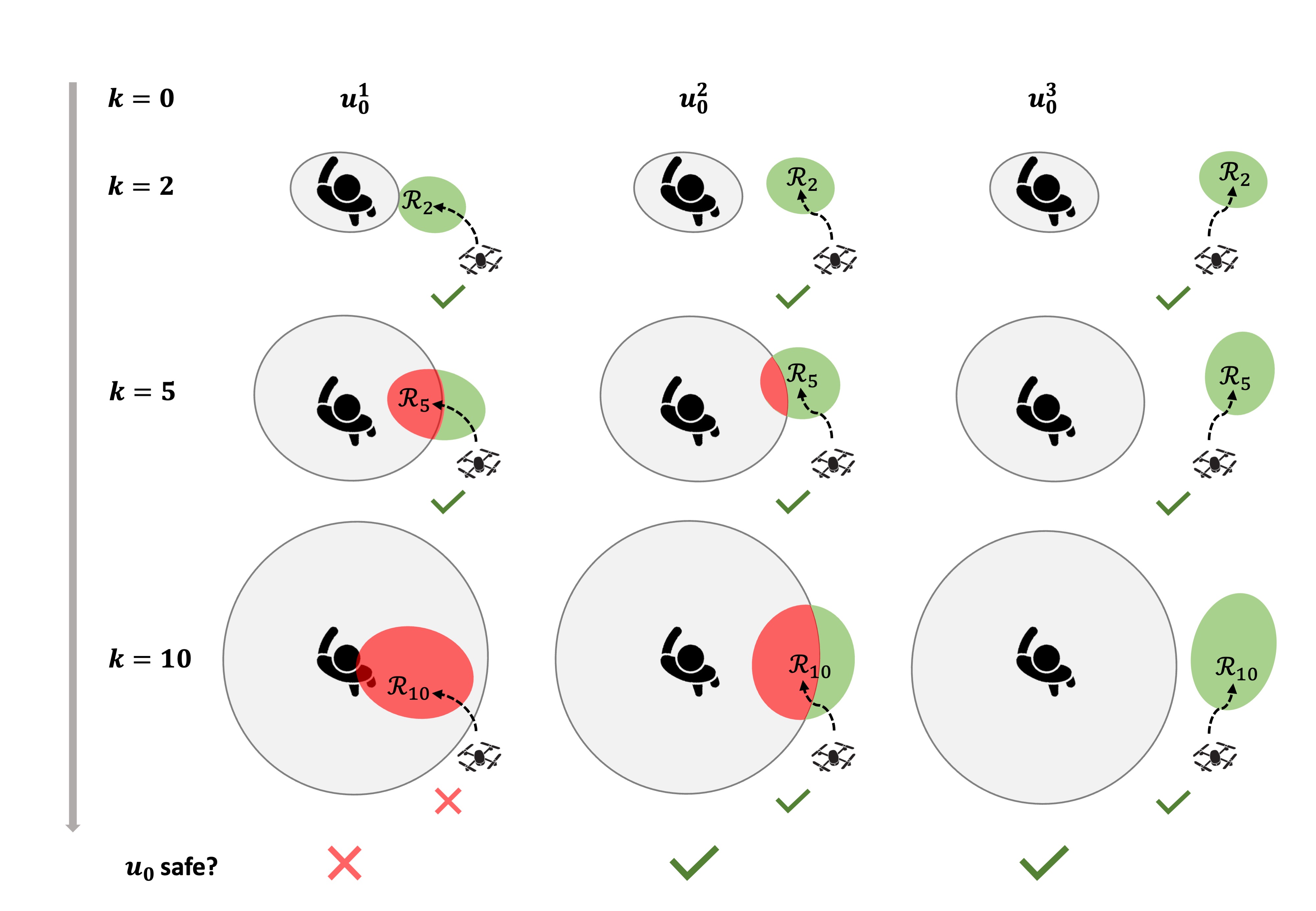}
\caption{Example evaluation of our safety constraint for three different initial control input choices. While the left choice leads to unsafe behavior, both the middle and the right choices for the initial control input are regarded safe as they leave options to avoid collision later on.}
\label{fig:reachability-constraint-overview}
\vspace{-0.3cm}
\end{figure}

We enforce~\eqref{eq:reachability_general} on each convex primitive separately. For $k>1$ this is a relaxation, necessary but not sufficient for non-containment of the \emph{union}, since the \ac{MAV} reachable set may lie inside the union while escaping each primitive on its own. These horizon terms therefore act as look-ahead viability conditions, not a certificate of a jointly collision-free future. At $k{=}1$, however, the reachable set is just the next state $\mathbf{x}_1=\mathbf{A}\mathbf{x}_0+\mathbf{B}\mathbf{u}_0$, as only $\mathbf{u}_0$ has acted, so exclusion from every primitive is exactly exclusion from their union, which is what Proposition 1 relies on.
\cref{fig:reachability-constraint-overview} illustrates the constraint for a single human set $\mathcal S_2$ and \ac{MAV} set $\mathcal S_1$, where enforcing $\mathcal S_1 \not\subseteq \mathcal S_2$ keeps an avoidance option available at that step. We next express \eqref{eq:reachability_general} as a constraint linear in $\mathbf{u}_0$, so that \eqref{eq:optimization_problem} is a convex \ac{QP}~\cite{boyd2004convex}.

\textbf{\ac{MAV} Reachable Set}
We represent the reachable \ac{MAV} \emph{positions} at step $k$, the state reachable set projected onto its position coordinates, as a zonotope
\begin{equation}
\mathcal{R}^R_k = \big\{\, \mathbf{y} \;\big|\; 
\mathbf{y} = \mathbf{c}^R_k + G_k \boldsymbol{\xi},\ \|\boldsymbol{\xi}\|_\infty \le 1 \big\},
\end{equation}
with center $\mathbf{c}^R_k$ and generator matrix $\mathbf{G}_k$. For the linear \ac{MAV} model~\eqref{eq:system_dynamics}, this exactly characterizes all positions reachable at step $k$, the generators encoding the directions and magnitudes of expansion.

The corresponding support function in the direction $\mathbf{n}_k$ is
\begin{equation}
\sup_{\mathbf{y}\in \mathcal{R}^R_k} \mathbf{n}_k^\top \mathbf{y}
= \mathbf{n}_k^\top \mathbf{c}^R_k + \|\mathbf{G}_k^\top \mathbf{n}_k\|_1,
\end{equation}
which is the exact support of the zonotope. Using the stacked dynamics~\eqref{eq:system_dynamics}, we separate the executed input $\mathbf{u}_0$ from the free future inputs $\mathbf{u}_{1:T-1}$ by partitioning $\mathbf{\Gamma}=[\,\mathbf{\Gamma}_1\ \ \mathbf{\Gamma}_f\,]$ into the block $\mathbf{\Gamma}_1$ acting on $\mathbf{u}_0$ and the block $\mathbf{\Gamma}_f$ acting on the free inputs, and write the latter as $\mathbf{u}_{1:T-1}=\mathbf{u}_{c,f} + E_{u,f} \boldsymbol{\xi}$ with $\mathbf{u}_{c,f}=\tfrac{1}{2}(\mathbf{u}_{\max}+\mathbf{u}_{\min})$, $E_{u,f}=\mathrm{diag}\big(\tfrac{1}{2}(\mathbf{u}_{\max}-\mathbf{u}_{\min})\big)$, and $\|\boldsymbol{\xi}\|_\infty\leq 1$. Writing $\mathbf{\Phi}_1$ and $\mathbf{\Gamma}_1$ for the blocks that map $\mathbf{x}_0$ and $\mathbf{u}_0$ to the \ac{MAV} position at step $k$, the reachable set conditioned on the executed input is
\begin{equation}
\mathcal{R}^R_k(\mathbf{u}_0) = \big\{\, \mathbf{c}^R_k + \mathbf{G}_k \boldsymbol{\xi} \;:\; \|\boldsymbol{\xi}\|_\infty \le 1 \big\},
\end{equation}
with center $\mathbf{c}^R_k = \mathbf{\Phi}_1 \mathbf{x}_0 + \mathbf{\Gamma}_1 \mathbf{u}_0 + \mathbf{\Gamma}_f \mathbf{u}_{c,f}$ and generators $\mathbf{G}_k = \mathbf{\Gamma}_f E_{u,f}$. The executed input thus shifts only the center, while the generators, and hence the support $\|\mathbf{G}_k^\top \mathbf{n}_k\|_1$, depend on the remaining control freedom alone.

\begin{figure}
    \centering
    \begin{tabular}{c c}
        \textbf{Complex} & \textbf{Simplified} \\
        \begin{subfigure}{0.45\linewidth}
            \centering
            \includegraphics[height=2.2cm]{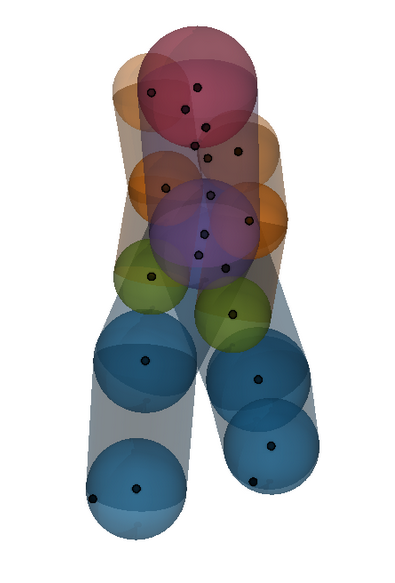}
            \caption{$\Delta t = 0.025s$}
        \end{subfigure} &
        \begin{subfigure}{0.45\linewidth}
            \centering
            \includegraphics[height=2.2cm]{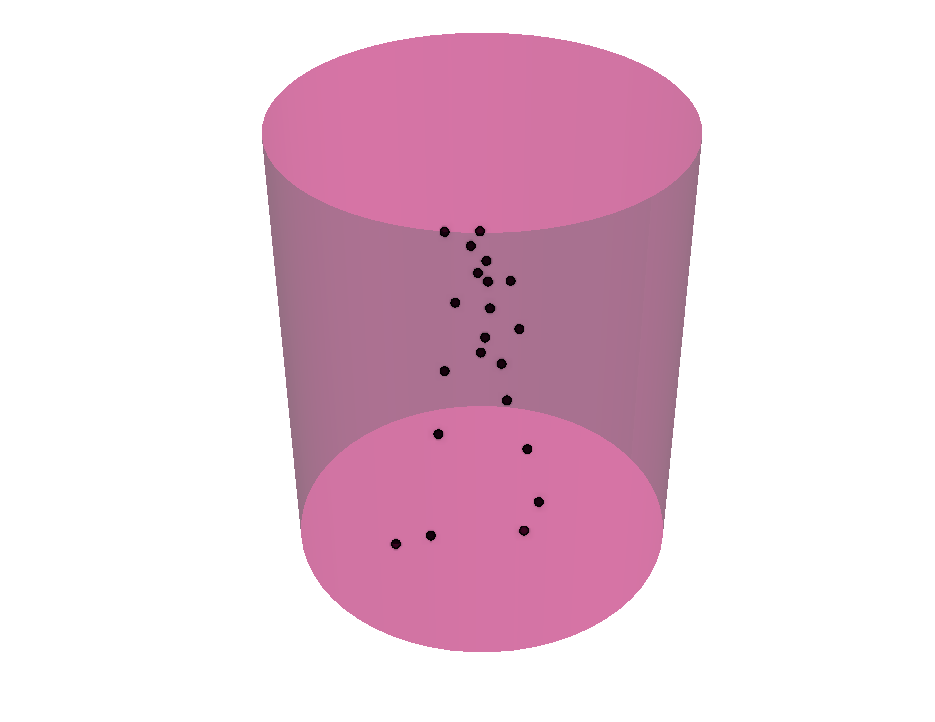}
            \caption{$\Delta t = 0.025s$}
        \end{subfigure} \\
        \begin{subfigure}{0.45\linewidth}
            \centering
            \includegraphics[height=2.2cm]{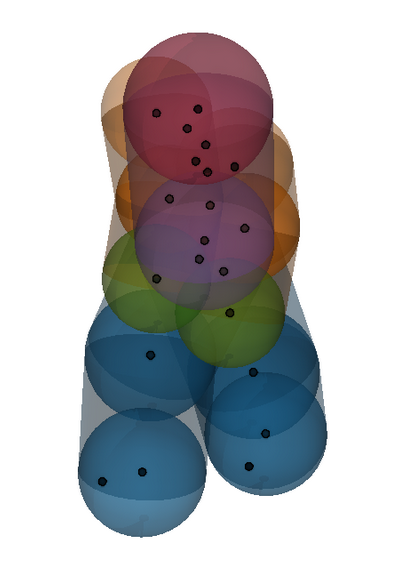}
            \caption{$\Delta t = 0.500s$}
        \end{subfigure} &
        \begin{subfigure}{0.45\linewidth}
            \centering
            \includegraphics[height=2.2cm]{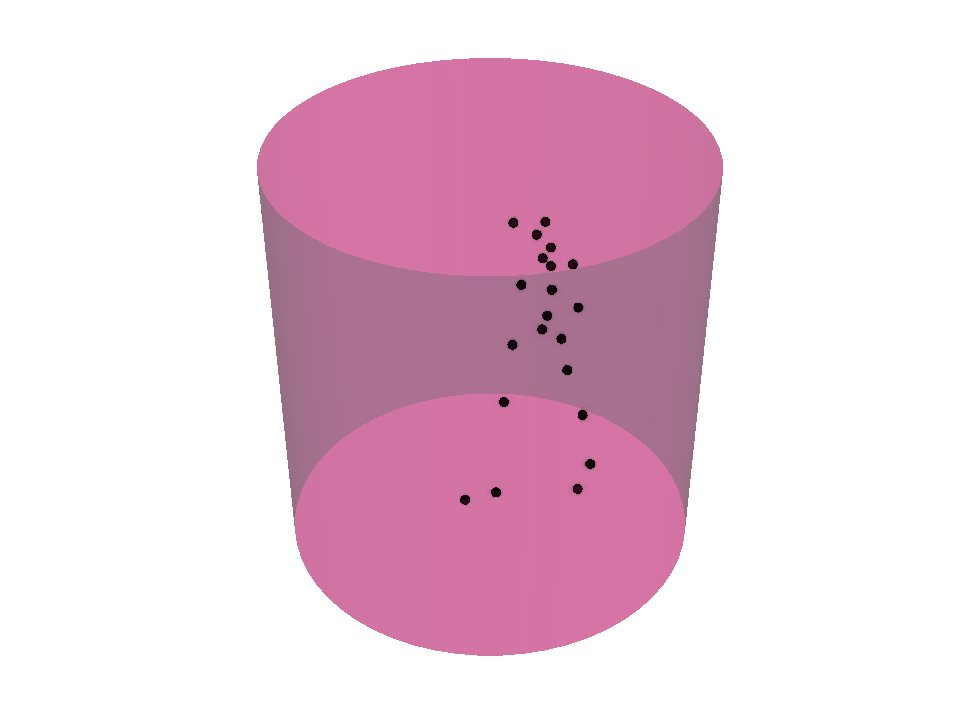}
            \caption{$\Delta t = 0.500s$}
        \end{subfigure} \\
    \end{tabular}
    \caption{Example constructions of the human reachable set for discretization steps $\Delta t = 0.025s$ and $\Delta t = 0.500s$. The black dots represent the 3D body joints at $t = 0s$.}
    \label{fig:hrs_examples}
    \vspace{-0.5cm}
\end{figure}

\textbf{Human Reachable Set}
\label{sec:human_reachable_set}The human reachable set (HRS) $\mathcal{R}^H_{k,i}$ combines two construction methods (\cref{fig:hrs_examples}).

The \emph{complex} set follows the body skeleton~\cite{schepp2022sara}, with capsules on torso and limb joints and spheres on head and hands whose radii expand over the horizon by a double-integrator model with bounded velocity and acceleration plus the robot size, giving 14 primitives per step. The \emph{simplified} set instead uses a single cylinder centered on the root's reachable trajectories, with radius set by their maximum deviation plus arm span and robot size. The simplified set over-approximates the admissible human occupancy over the whole horizon, whereas the complex set is tight near-term but ceases to over-approximate it long-term. Each convex primitive is one constraint $h_i$, so $i$ indexes all primitives of all humans, 14 per human for the \emph{complex} set and one for the \emph{simplified}. We therefore combine them, using \emph{complex} before a switch time $t_s = 0.2s$ for near-term tightness and \emph{simplified} afterwards to stay conservative (Assumption 1) while cutting the constraint count (\cref{sec:experiments}). Unlike static 3D bounding boxes~\cite{song2024mpcobstacles}, these sets capture time-varying full-body motion, embed in convex optimization, and can be evaluated efficiently online.

These sets grow with the horizon and the motion bounds, whereas $\mathcal{R}^R_k$ is computed \emph{exactly} under the linear dynamics and control limits, an asymmetry, conservative for humans and exact for the \ac{MAV}, that is what ensures safety.

\textbf{Constraint Definition}
Let $\mathbf{m}_k \in \partial \mathcal{R}^H_{k,i}$ be the boundary point of the convex primitive $\mathcal{R}^H_{k,i}$ closest to $\mathcal{R}^R_k$, and $\mathbf{n}_k$ the corresponding outward normal, both evaluated at the nominal $\mathcal{R}^R_k$ and held fixed during the solve. Because the executed input $\mathbf{u}_0$ only translates $\mathcal{R}^R_k$ and leaves its generators unchanged, keeping $\mathcal{R}^R_k$ on the far side of this fixed supporting hyperplane is a sufficient condition for non-containment,
\begin{equation}
\sup_{\mathbf{y} \in \mathcal{R}^R_k} \mathbf{n}_k^\top (\mathbf{y} - \mathbf{m}_k) \geq \varepsilon,
\end{equation}
for some $\varepsilon > 0$. Substituting the conditioned center $\mathbf{c}^R_k$ and its support $\|\mathbf{G}_k^\top \mathbf{n}_k\|_1$ and collecting the terms independent of $\mathbf{u}_0$ into
\begin{equation}
b_k = \varepsilon - \|\mathbf{G}_k^\top \mathbf{n}_k\|_1 + \mathbf{n}_k^\top\!\left(\mathbf{m}_k - \mathbf{\Phi}_1 \mathbf{x}_0 - \mathbf{\Gamma}_f \mathbf{u}_{c,f}\right),
\end{equation}
this condition becomes linear in the executed input $\mathbf{u}_0$:
\begin{equation}
(\mathbf{n}_k^\top \mathbf{\Gamma}_1)\, \mathbf{u}_0 \geq b_k .
\label{eq:safety_constraint}
\end{equation}

The constraint in \eqref{eq:safety_constraint} is linear and efficient to compute, and under the conditions formalized next it recursively assures safety in an MPC-like control scheme.
If \eqref{eq:optimization_problem} is infeasible, e.g.\ after a perception failure, we solve a softened fallback that relaxes the constraints to $h_i \geq -s_{i,k}$ and minimizes $\lambda\sum_{i,k} s_{i,k}$ and distances the \ac{MAV} from the human.

 \subsection{Recursive Safety Guarantee}
\label{sec:safety_analysis}

We now show that \eqref{eq:safety_constraint} preserves safety for all future time, by a receding-horizon recursive-feasibility argument in the spirit of predictive safety filters~\cite{wabersich2018safetyfilter}.
Index the executed closed-loop steps by $t$. At each $t$ we observe $\mathbf{x}(t)$ and the joints $\mathbf{J}^H(t)$, build the human reachable sets $\mathcal{R}^H_{k\mid t}$ ($k$ steps ahead of $t$), solve \eqref{eq:optimization_problem} over $[t,t{+}T]$, and execute the first input $\mathbf{u}_0^\star(t)$. Let $\mathcal{K}(t)$ be the true human occupancy at $t$ inflated by $\mathrm{R}_{\text{MAV}}$ and the margin, so clearance is kept iff $\pos{W}{B}(t)\notin\mathcal{K}(t)$.

\smallskip
\textbf{Assumption 1 (Conservative human prediction).} The set $\mathcal{R}^H_{k\mid t}$ over-approximates \emph{every} position any body part can occupy $k$ steps after $t$ that is consistent with the velocity and acceleration bounds $v_{i,\max},a_{i,\max}$ used to build it (Table~\ref{tab:parameters}). For motions that respect these velocity and acceleration bounds, the true inflated occupancy is contained, $\mathcal{K}(t{+}k)\subseteq\mathcal{R}^H_{k\mid t}$, and re-observation only refines the prediction, $\mathcal{R}^H_{k\mid t+1}\subseteq\mathcal{R}^H_{k+1\mid t}$, for all $t,k$.

\smallskip
\textbf{Assumption 2 (Exact model and observation).} The executed \ac{MAV} follows the identified linear dynamics $\mathbf{x}_{k+1}=\mathbf{A}\mathbf{x}_k+\mathbf{B}\mathbf{u}_k$ within the control box $\mathcal{U}$ (Sec.~\ref{sec:preliminaries}), and the state $\mathbf{x}(t)$ and joints $\mathbf{J}^H(t)$ are available without error at every replan.

\smallskip
\textbf{Assumption 3 (Recursive feasibility).} Whenever the hard-constrained problem~\eqref{eq:optimization_problem} is feasible at $t$, executing its optimized first input $\mathbf{u}_0^\star(t)$ leads to a state $\mathbf{x}(t{+}1)$ for which, after re-observing the human and recomputing the supporting hyperplanes, \eqref{eq:optimization_problem} is again feasible at $t{+}1$. This is a property of the closed loop of the implemented receding-horizon controller, not one implied by the reachability constraint alone: our constraint enforces one-step non-containment but adds no terminal-invariance condition, so recursive feasibility is a separate property, which we assess empirically (\cref{sec:experiments}).

\smallskip
\textbf{Proposition 1 (One-step sampled safety).} Under Assumptions 1 and 2, whenever~\eqref{eq:optimization_problem} is feasible at $t$ the executed input $\mathbf{u}_0^\star(t)$ keeps the next step clear, $\pos{W}{B}(t{+}1)\notin\mathcal{K}(t{+}1)$.

\smallskip
\textbf{Proof.} The optimizer returns an input $\mathbf{u}_0^\star$ satisfying~\eqref{eq:safety_constraint}, which by construction implies the non-containment~\eqref{eq:reachability_general}. Its $k{=}1$ term fixes a single point, the executed next position $\pos{W}{B}(t{+}1)$ from $\mathbf{x}(t{+}1)=\mathbf{A}\mathbf{x}(t)+\mathbf{B}\mathbf{u}_0^\star$, so non-containment gives $\pos{W}{B}(t{+}1)\notin\mathcal{R}^H_{1\mid t}$, and $\mathcal{R}^H_{1\mid t}\supseteq\mathcal{K}(t{+}1)$ by Assumption 1 gives $\pos{W}{B}(t{+}1)\notin\mathcal{K}(t{+}1)$. This step rests on Assumptions 1--2 alone.

\smallskip
\textbf{Corollary 1 (Conditional recursive safety).} If, in addition, the initial state is collision-free, $\pos{W}{B}(0)\notin\mathcal{K}(0)$, \eqref{eq:optimization_problem} is feasible initially, and the closed loop is recursively feasible (Assumption 3), then \eqref{eq:optimization_problem} stays feasible at every replanning instant and the closed loop is collision-free for all $t\geq0$, $\pos{W}{B}(t)\notin\mathcal{K}(t)$.

\smallskip
\textbf{Proof.} Induction on $t$. At $t{=}0$ the problem is feasible and the state collision-free by hypothesis. Whenever \eqref{eq:optimization_problem} is feasible at $t$, Proposition 1 gives $\pos{W}{B}(t{+}1)\notin\mathcal{K}(t{+}1)$, and Assumption 3 leaves \eqref{eq:optimization_problem} feasible at $t{+}1$, so the same argument applies at the next replanning instant. Induction yields feasibility, hence clearance, for all $t\geq0$.

\smallskip
Assumption 3 is the recursive-feasibility condition required by the receding-horizon argument. One could enforce it by constraining the successor state to a robust control-invariant safe set of a verified backup controller, but computing such a set for the coupled worst-case human--\ac{MAV} game is an \ac{HJ} reachability problem whose cost grows exponentially with the state dimension. For 3D \ac{MAV} motion against one or several full-body humans it is therefore intractable, which is precisely the offline computation our formulation is designed to avoid. We instead state recursive feasibility as an assumption and assess it empirically: as reported in \cref{sec:experiments}, the hard-constrained problem stays feasible on all nominal runs. Proposition 1 gives one-step sampled safety under Assumptions 1--2 whenever the problem is feasible, and Assumption 3 lets Corollary 1 apply it recursively, yielding the closed-loop guarantee. The guarantee is nominal. Perception noise and model mismatch are handled in practice by the conservative over-approximation, the margin $\mathrm{R}_{\text{MAV}}$, and reachable-set inflation over unobserved intervals. 

\subsection{Static Obstacles}
\label{sec:workspace}
Indoor and urban deployments also contain static structure such as walls, floors, and ceilings. We model the admissible workspace as an intersection of half-spaces, one per static surface $j$ with inward normal $\mathbf{n}_j$ and offset $b_j$, and define the obstacle-avoidance function
\begin{equation}
f_j(\mathbf{x}_k) = \mathbf{n}_j^\top\, \pos{W}{B} - b_j - \delta,
\label{eq:workspace_constraint}
\end{equation}
the signed distance from the \ac{MAV} position $\pos{W}{B}$ to surface $j$ with safety margin $\delta$. Since the position is a linear function of the state, and hence of the inputs through \eqref{eq:system_dynamics}, each $f_j(\mathbf{x}_k)\geq 0$ is a linear constraint that enters \eqref{eq:optimization_problem} directly and, unlike the softenable human constraint, is kept hard. As shown in \cref{tab:constrained}, this lets the \ac{MAV} navigate constrained corridors where a simple flyover is impossible, remaining safe with respect to both the human and the convex workspace, a setting the human-only baselines cannot represent.

%% file: sections/06_experiments.tex
\section{Experiments}
\label{sec:experiments}
We evaluate our method on two tasks, setpoint navigation and visual servoing for human tracking, in both simulation and real-world environments. The control framework is implemented in C++ using the HPIPM solver~\cite{frison2020hpipm} and runs onboard at \SI{40}{\hertz}.

\begin{table}[H]
\centering
\begin{tabular}{cc|cc|cc}
\toprule
\textbf{Parameter} & \textbf{Value} & \textbf{Parameter} & \textbf{Value} & \textbf{Parameter} & \textbf{Value} \\
\midrule
$\rho_{Head}$ & 0.2 & $\rho_{Torso}$ & 0.3 & $\rho_{Arm}$ & 0.205 \\
$\rho_{Hand}$ & 0.1 & $v_{i, \max}$ & 1.0 & $a_{i, \max}$ & 1.0 \\
$\lambda$ & 1000 & $R_{\text{MAV}}$ & 0.5 & $\tau_{\min}$ & -0.5 \\
$\tau_{\max}$ & 0.25 & $\theta_{\min}$ & $ - \ang{15}$ & $\theta_{\max}$ & $\ang{15}$ \\
$\phi_{\min}$ & $ - \ang{15}$ & $\phi_{\max}$ & $\ang{15}$ & $b_1$ & 1.0 \\
$b_2$ & 0.1 & $b_3$ & 1.0 & $b_4$ & 0.1 \\
$c_i$ & 0.01 & $T_s$ & 0.025s & $t_s$ & 0.2s \\
\bottomrule
\end{tabular}
\caption{Parameters for reachable sets, box constraints, and \ac{MAV} model in simulation.}
\label{tab:parameters}
\end{table}

\subsection{Simulation Experiments}
\label{sec:simulation_exp}
Experiments are conducted in simulation using Ignition Gazebo combined with PX4 to reproduce realistic flight dynamics, using the parameters shown in \cref{tab:parameters}.
As commonly done in the social navigation literature, we evaluate on goal-directed navigation, where the \ac{MAV} plans and executes a trajectory from an initial state to a goal position while maintaining safety around surrounding humans. Real human body motions serve as dynamic obstacles. We replay 50 motion sequences from the AMASS~\cite{naureen2019amass} dataset as ground-truth human motion, and sample initial states and goals uniformly from the feasible regions. Unless noted, we use a horizon of 40 steps at a temporal resolution of \SI{0.025}{\second}.

The parameterization of the human reachable set builds on prior human motion analysis~\cite{schepp2022sara, giannoulaki2024walkingspeed} and is validated on AMASS~\cite{naureen2019amass}, where the resulting sets cover \SI{100}{\percent} of observed body-joint positions over the horizon. Body parts are modeled as second-order systems with initial radii $\rho_i$ and bounds $v_{i,\max}$, $a_{i,\max}$, plus the \ac{MAV} radius $R_{\text{MAV}}$. The weight $\lambda$ penalizes constraint violation in the softened fallback QP that runs only when the nominal problem is infeasible. Because it merely restores feasibility rather than shaping the nominal trajectory, the method is insensitive to its exact value, and we keep $\lambda{=}1000$ throughout.

\begin{table*}[t!]
\centering
\setlength{\tabcolsep}{3pt}
\renewcommand{\arraystretch}{1.15}
\begin{tabular}{l cccc cccc}
\toprule
\textbf{Method} &
\multicolumn{4}{c}{\textbf{1 Human}} &
\multicolumn{4}{c}{\textbf{2 Humans}} \\
\cmidrule(lr){2-5} \cmidrule(lr){6-9}
& \makecell{\textbf{Coll. Avoid.}\\$\uparrow$ [\%]}
& \makecell{\textbf{SR}\\$\uparrow$ [\%]}
& \makecell{\textbf{TTG}\\$\downarrow$ [s]}
& \makecell{\textbf{Solver}\\$\downarrow$ [ms]}
& \makecell{\textbf{Coll. Avoid.}\\$\uparrow$ [\%]}
& \makecell{\textbf{SR}\\$\uparrow$ [\%]}
& \makecell{\textbf{TTG}\\$\downarrow$ [s]}
& \makecell{\textbf{Solver}\\$\downarrow$ [ms]} \\
\midrule
No Constraint             & 94/92   & 94/92   & 3.96/3.96 & \textbf{0.05} & 89/91   & 89/91   & 3.56/3.56 & \textbf{0.07} \\
3D-ORCA~\cite{vandenberg2011orca}          & 98/96   & 84/80   & 4.98/4.87 & 0.05 & 100/93  & 91/87   & 4.74/4.74 & 0.07 \\
CBF-QP~\cite{ames2017cbf}                  & 100/100 & 100/98  & 4.43/4.51 & 0.06 & 100/98  & 98/91   & 4.11/3.99 & 0.11 \\
MPC-CBF~\cite{zeng2021mpccbf}              & 100/96  & 100/96  & 4.17/4.14 & 0.57 & 100/91  & 100/91  & 3.94/3.82 & 1.39 \\
Salazar et al.~\cite{salazar2024hierarchical} & 96/90 & 82/78 & 4.91/5.03 & 0.06 & 98/98 & 91/89 & 4.72/4.61 & 0.08 \\
\midrule[0.05pt]
DC + Static Body          & 94/90   & 94/90   & 4.15/4.17 & 10.80 & 91/89   & 91/89   & 3.68/3.86 & 21.44 \\
DC + Const. Velocity      & 94/92   & 94/92   & 4.10/4.19 & 10.78 & 91/89   & 91/89   & 3.74/3.75 & 21.46 \\
DC + Forecast             & 94/90   & 94/90   & 4.09/4.12 & 10.77 & 91/89   & 91/89   & 3.74/3.74 & 21.19 \\
\midrule[0.05pt]
RC + Simplified HRS       & 100/100 & 98/94   & 4.19/4.26 & 2.90 & 100/100 & 96/91   & 3.84/3.78 & 5.46 \\
\textbf{Ours (RC + Hybrid HRS)} & \textbf{100/100} & \textbf{98/94} & \textbf{4.17/4.19} & 3.90 & \textbf{100/100} & \textbf{96/93} & \textbf{3.83/3.87} & 8.40 \\
\bottomrule
\end{tabular}
\caption{Simulation evaluation. Each cell reports \emph{det/noisy}, i.e.\ under a perfect forecast and under an imperfect one where the executed human motion deviates from the prediction by a seeded per-sequence velocity drift (std \SI{0.6}{\meter\per\second}). \emph{SR} is success rate and \emph{TTG} time-to-goal. With a perfect forecast all safety-aware methods are comparable, but under prediction error only the reachability variants stay fully collision-free while also reaching the goal faster than the external safe baselines, carrying a formal guarantee (Proposition 1) and, unlike prior baselines, extending to static obstacles (\cref{tab:constrained}).}
\label{tab:reachability}
\vspace{-0.2cm}
\end{table*}

\textbf{Metrics.}
\emph{Collision Avoidance} is the fraction of runs whose minimum \ac{MAV}-to-body distance stays $\geq\SI{0.5}{\meter}$. \emph{Success Rate} additionally requires reaching the goal within the time budget, whose elapsed time is the \emph{Time-To-Goal}. \emph{Solver Time} covers constraint and reachable-set assembly plus the solve. For the constrained workspace, \emph{Obstacle Avoidance} is the fraction of runs not colliding with the static scene.

\textbf{Baselines.}
We compare along three axes, namely external safe control, the constraint formulation, and the reachable-set representation. As external references we implement \emph{3D-ORCA}~\cite{vandenberg2011orca} (humans as moving spheres on the 2D root), the \emph{CBF-QP} safety filter~\cite{ames2017cbf}, and \emph{MPC-CBF}~\cite{zeng2021mpccbf} (a discrete-time barrier over the horizon). All three receive the same body primitives, human input, and low-level model, isolating the constraint. We further include the hierarchical controller of~\cite{salazar2024hierarchical} in its native 2D form, and a \emph{No Constraint} reference. The distance-constraint (DC) baselines enforce clearance to a \emph{predicted, non-inflated} body under static, constant-velocity, or forecast motion models. Finally, we ablate the human reachable set, with \emph{Simplified} using only the cylinder and \emph{Ours} the hybrid (complex short-term, simplified long-term).

\textbf{Free-space navigation.}
\cref{tab:reachability} reports both regimes. With a perfect forecast the strongest methods reach \SI{100}{\percent} collision avoidance with comparable success rates, so our safety constraint costs little efficiency. The \emph{No Constraint} reference, the DC baselines, and the hierarchical controller~\cite{salazar2024hierarchical} already fall short of full avoidance, the latter slowest to the goal. Under imperfect prediction, every barrier-, prediction-, and heuristic-based baseline drops further below \SI{100}{\percent} (CBF-QP to 98\%, MPC-CBF to 91\%, Salazar to 90\%, DC to 89\%), while both reachability variants stay fully collision-free, as the worst-case set absorbs the forecast error their constraints cannot. Beyond this margin, our constraint carries a \emph{formal} safety guarantee that prior baselines lack and extends to static obstacles, observation delays are handled in practice by inflating the reachable sets. Among the method variants hybrid set matches the simplified-only variant here and outperforms the complex-only set in the corridor (82\% vs.\ 64\%, \cref{tab:constrained}), justifying the combination. It also reaches the goal up to $\sim$\SI{0.9}{\second} sooner than 3D-ORCA and the hierarchical baseline. Notably, 3D-ORCA reasons only about the planar root and, ignoring the full-body extent, drops to 93--96\% avoidance, underscoring the value of the complete 3D body. Across our simulation and real-world runs the hard-constrained \ac{QP} stayed feasible in all nominal cases, empirically supporting Assumption 3, the softened fallback engaged only in degenerate ones, gross perception failures such as a body detection jumping, or ill-posed starts such as the \ac{MAV} initialized right next to a human. Even then no collision occurred, as softening keeps driving the \ac{MAV} away from the human rather than abandoning avoidance.

\begin{table}[ht!]
\centering
\setlength{\tabcolsep}{4pt}
\renewcommand{\arraystretch}{1.2}
\begin{tabular}{l ccccc}
\toprule
\textbf{Method}
& \makecell{\textbf{Human}\\\textbf{Avoid. $\uparrow$}}
& \makecell{\textbf{Obstacle}\\\textbf{Avoid. $\uparrow$}}
& \makecell{\textbf{Success}\\\textbf{Rate $\uparrow$}}
& \makecell{\textbf{Min.}\\\textbf{Dist. $\uparrow$}}
& \makecell{\textbf{Solver}\\\textbf{Time $\downarrow$}} \\
& [\%] & [\%] & [\%] & [m] & [ms] \\
\midrule
No Constraint            & 14  & 100 & 14  & 0.21 & 0.18 \\
3D-ORCA~\cite{vandenberg2011orca}         & 72  & 100 & 26  & 0.75 & \textbf{0.06} \\
CBF-QP~\cite{ames2017cbf}                 & 94  & 94  & 50  & 1.09 & 0.08 \\
MPC-CBF~\cite{zeng2021mpccbf}             & 62  & 96  & 56  & 0.77 & 0.76 \\
\midrule[0.05pt]
Complex HRS         & 84  & 100 & 64  & 0.97 & 3.40 \\
\textbf{Ours}       & \textbf{100} & \textbf{100} & \textbf{82} & \textbf{1.41} & 4.02 \\
\bottomrule
\end{tabular}
\caption{Constrained-workspace evaluation: a \SI{2}{\meter}-wide corridor with a low ceiling below the human's vertical reach, so a flyover is impossible. Only our method stays simultaneously safe w.r.t.\ humans \emph{and} obstacles (both \SI{100}{\percent}) while completing the task most often.}
\label{tab:constrained}
\vspace{-0.3cm}
\end{table}

\textbf{Constrained workspace and static obstacles.}
\label{sec:constrained_exp}
To assess navigation among static obstacles, we place the human on the path in the corridor of \cref{tab:constrained}, where a low ceiling makes a flyover impossible. Only our method is simultaneously safe w.r.t.\ the human and the workspace (both \SI{100}{\percent}) while completing the task most often (\SI{82}{\percent}) and keeping the largest clearance. Every reactive baseline fails at least one requirement. MPC-CBF and 3D-ORCA collide with the human in 38\% and 28\% of runs, and CBF-QP, though largely safe, reaches the goal only half the time. The Complex-set variant matches our workspace safety but still collides in 16\% of runs, as the complex set no longer over-approximates the human occupancy late in the horizon and forfeits the guarantee there. The hybrid set stays conservative throughout and improves both safety and completion.

\begin{table}[H]
\centering
\begin{tabular}{lcccc}
\toprule
\textbf{Horizon}
& \makecell{\textbf{Coll.}\\\textbf{Avoid. $\uparrow$ [\%]}}
& \makecell{\textbf{Success}\\\textbf{Rate $\uparrow$ [\%]}}
& \makecell{\textbf{Time To}\\\textbf{Goal $\downarrow$ [s]}}
& \makecell{\textbf{Solver}\\\textbf{Time $\downarrow$ [ms]}} \\
\midrule
20 & 100 & 84 & 7.53 & 1.09 \\
40 & 100 & 98 & 4.17 & 3.90 \\
60 & 100 & 98 & 3.52 & 9.05 \\
\bottomrule
\end{tabular}
\caption{Effect of horizon length in the single-human scene. Safety holds at every horizon. A short horizon yields slow, conservative trajectories, while extending the horizon reduces time-to-goal only up to a point, after which it mainly adds computation. A horizon of 40 balances goal-reaching speed and solver time.}
\label{tab:horizon_length}
\vspace{-0.3cm}
\end{table}

\textbf{Horizon length.}
The horizon governs the lookahead available to the reachability recursion (\cref{tab:horizon_length}), and safety is maintained at every tested horizon. A shorter horizon yields slow, conservative trajectories (time-to-goal \SI{7.5}{\second} at 20 steps vs.\ \SI{4.2}{\second} at 40), while extending it to 60 saves only \SI{0.65}{\second} more yet doubles the solver time, so 40 is the best balance.

\subsection{Real-World Experiments}
\label{sec:simulation_real}
We validate transfer to an embedded \ac{MAV}, a quadrotor with an Intel RealSense D455 and a Jetson Orin NX running the full stack, comprising state estimation with OKVIS2~\cite{leutenegger2022okvis2}, 2D human tracking lifted to 3D joints, our controller, and actuation through PX4~\cite{px42015meier}. Onboard deployment requires real-time perception, yet monocular 3D pose estimators either estimate the global root poorly~\cite{kanazawa2017hmr} or rely on heavy optimization unsuited to embedded platforms~\cite{schaefer2023glopro}. To stress-test our controller we deliberately adopt a lightweight, noise-prone pipeline, detecting humans with a real-time 2D pose model~\cite{bin2018bodytracking}, back-projecting 3D joints onto synchronized depth, and associating by pixel proximity, unfiltered. Human motion forecasts, produced by MotionMixer~\cite{bouazizi2022motionmixer} from the last 10 observed frames, enter only the objective. The safety constraint does not depend on them, and the near-identical DC baselines (\cref{tab:reachability}) confirm that placing a predictor in the constraint does not by itself confer safety. Missed detections and latency are handled by growing each joint's reachable set with its elapsed unobserved time, folding latency into the safety reasoning.

\textbf{Runtime.} On the Jetson Orin NX, solver assembly and solve take $\approx\SI{14}{\milli\second}$ (reachable-set assembly $\approx\SI{6}{\milli\second}$), enabling re-planning at the \SI{40}{\hertz} odometry rate alongside the perception stack ($\approx\SI{15}{\hertz}$).

\textbf{Results.} We evaluate visual servoing, where the \ac{MAV} must keep facing the front of a freely moving human at \SI{3}{\meter} (\cref{fig:real_world_experiments}). Across all 8 trials (each $\approx$1 min), with the human walking and turning about the space, the constraint preserved safety in every run without intervention. The \ac{MAV} kept an average distance of \SI{3.2}{\meter} and stayed within 20\% of the \SI{3}{\meter} reference $\approx$85\% of the time. Reachable-set inflation on stale joints kept it safe through the tracker's frequent dropouts and depth outliers, none of which are covered by the nominal assumptions.

%% file: sections/07_conclusion.tex
\section{Conclusion}
\label{sec:conclusion}
We proposed an \ac{MPC} framework whose linear reachability constraint provides a formal recursive safety guarantee, under stated assumptions, while retaining a \ac{QP} formulation.
Unlike \ac{HJ} approaches, it avoids heavy precomputation and model simplification, and accommodates resource-constrained platforms, slow updates, and static workspace constraints in real time. Using it, we are, to our knowledge, the first to demonstrate safe yet effective \ac{MAV} navigation in close proximity to moving humans in full 3D.
Experiments showed that, unlike prior barrier- and prediction-based baselines, our method preserves safety under imperfect prediction and in constrained workspaces at little cost, matching the best baselines under a perfect forecast. By reducing a reachability-based safety guarantee to linear constraints on the executed input, our framework runs onboard while reasoning over the full 3D human body.

The human-motion bounds $v_{i,\max}$ and $a_{i,\max}$ that size the reachable set are a tunable safety parameter, set conservatively from pedestrian-motion studies~\cite{schepp2022sara, giannoulaki2024walkingspeed} as well as empirical evaluations on AMASS~\cite{naureen2019amass}. Raising them extends coverage to faster motion at the cost of efficiency. Our evaluation also uses replayed, non-reactive human motion, so reactive and adversarial crowds remain open, alongside tighter, learned reachable sets that adapt these bounds online.